\documentclass[letterpaper]{article} 
\usepackage[submission]{aaai2026}  
\usepackage{times}  
\usepackage{helvet}  
\usepackage{courier}  
\usepackage[hyphens]{url}  
\usepackage{graphicx} 
\usepackage{natbib}  
\usepackage{caption} 
\usepackage{amsfonts} 
\usepackage{algorithm}
\usepackage{algorithmic}

\usepackage{booktabs}
\usepackage{multirow}
\usepackage{amsmath}
\usepackage{tabularx}

\usepackage{newfloat}
\usepackage{listings}
\DeclareCaptionStyle{ruled}{labelfont=normalfont,labelsep=colon,strut=off} 
\floatstyle{ruled}
\newfloat{listing}{tb}{lst}{}
\floatname{listing}{Listing}
\def\showauthors@on{T}
\title{Enhancing Molecular Property Prediction\\with Knowledge from Large Language Models}
\author{
    Peng Zhou\textsuperscript{\rm 1, \rm 2},
    Lai Hou Tim\textsuperscript{\rm 2},
    Zhixiang Cheng\textsuperscript{\rm 1},
    Kun Xie\textsuperscript{\rm 2, \rm 3}, \\
    Chaoyi Li\textsuperscript{\rm 1},
    Wei Liu\textsuperscript{\rm 2}, 
    Xiangxiang Zeng\textsuperscript{\rm 1}
}
\affiliations{
    \textsuperscript{\rm 1}Hunan University\\
    \textsuperscript{\rm 2}Tencent AI for Life Science Lab\\
    \textsuperscript{\rm 3}The Chinese University of Hong Kong\\
}

\usepackage{bibentry}

\begin{document}

\maketitle

\begin{abstract}
Predicting molecular properties is a critical component of drug discovery. Recent advances in deep learning—particularly Graph Neural Networks (GNNs)—have enabled end-to-end learning from molecular structures, reducing reliance on manual feature engineering. However, while GNNs and self-supervised learning approaches have advanced molecular property prediction (MPP), the integration of human prior knowledge remains indispensable, as evidenced by recent methods that leverage large language models (LLMs) for knowledge extraction. Despite their strengths, LLMs are constrained by knowledge gaps and hallucinations, particularly for less-studied molecular properties. In this work, we propose a novel framework that, for the first time, integrates knowledge extracted from LLMs with structural features derived from pre-trained molecular models to enhance MPP. Our approach prompts LLMs to generate both domain-relevant knowledge and executable code for molecular vectorization, producing knowledge-based features that are subsequently fused with structural representations. We employ three state-of-the-art LLMs—GPT-4o, GPT-4.1, and DeepSeek-R1—for knowledge extraction. Extensive experiments demonstrate that our integrated method outperforms existing approaches, confirming that the combination of LLM-derived knowledge and structural information provides a robust and effective solution for MPP.
\end{abstract}

\begin{figure*}[t]
  \centering
  \includegraphics[width=1\textwidth]{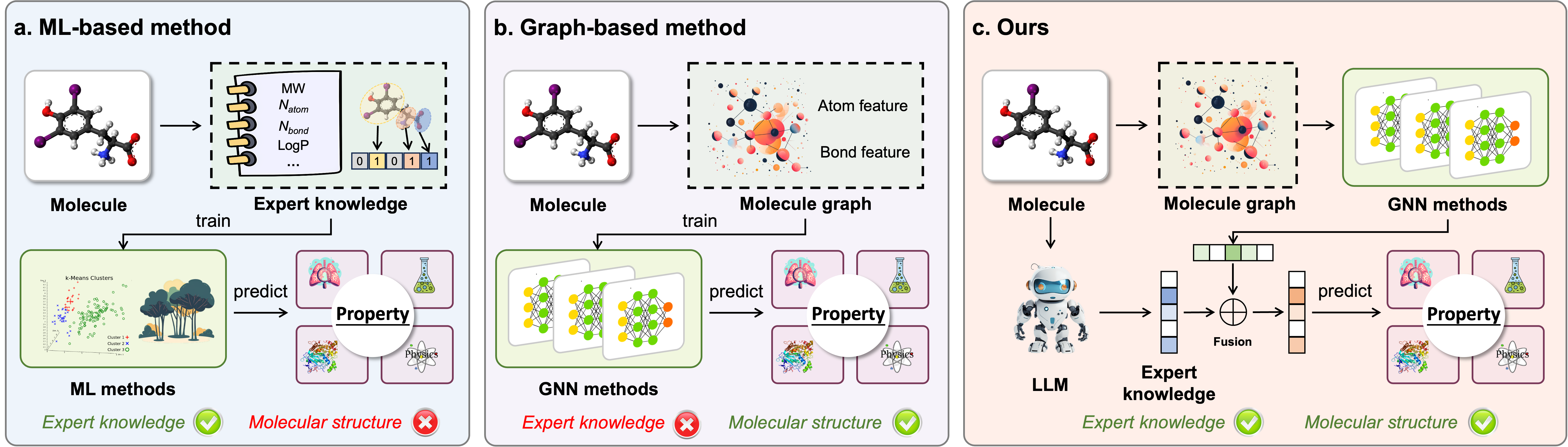}
  \caption{(a) Traditional machine learning algorithms rely heavily on expert knowledge and feature engineering, limiting their adaptability and generalization. (b) Graph neural networks, while directly mapping molecular graphs to properties, often ignore expert insights. (c) We bridges these gaps by using LLMs to incorporate expert knowledge with graph structures, enhancing the prediction of molecular properties.}
  \label{fig:framework}
\end{figure*}

\section{Introduction}
Predicting molecular properties is an essential task in the drug discovery process. Computational approaches to MPP not only accelerate drug screening but also improve cost efficiency. Traditional computational methods for molecular property prediction (MPP) typically involve extracting molecular fingerprints or carefully engineered features, followed by the application of machine learning algorithms such as Support Vector Machines (SVM) \citep{cortes1995support} and Random Forests (RF) \citep{ho1998random}. However, these methods heavily depend on domain experts for feature extraction and are susceptible to human knowledge biases \citep{merkwirth2005automatic, degen2008art}. The advent of deep learning has partially addressed these limitations. Deep learning approaches can better leverage the growing availability of data and are less reliant on manual feature engineering. In particular, the application of Graph Neural Networks (GNNs) in the molecular domain has shown significant promise, as molecules can naturally be represented as graph structures. These models can be trained end-to-end directly on molecular graphs, enabling them to capture higher-order nonlinear relationships more effectively, eliminate human biases, and dynamically adapt to different tasks \cite{wu2020comprehensive, mayr2018large, wieder2020compact}.

In recent years, numerous GNN-based methods have been proposed, many of which focus on developing improved self-supervised learning techniques \citep{liu2021self}, leveraging contrastive learning \citep{khosla2020supervised}, and deriving more robust molecular representations from large volumes of unlabeled molecular data \citep{liu2022pre, wang2022molecular}. However, we contend that human prior knowledge will remain indispensable for the foreseeable future. For example, by explicitly constraining chemical bonds between molecules and targets, Interformer \citep{lai2024interformer} can predict more plausible chemical conformations. Similarly, the incorporation of multiple sequence alignments enables AlphaFold to more accurately predict molecular and protein conformations \citep{jumper2021highly}. More relevantly, LLM4SD \citep{zheng2025large} leverages LLMs to extract human knowledge for molecular vectorization, subsequently employing Random Forests for MPP, and outperforms GNN-based methods on several tasks. LLMs such as ChatGPT \citep{openaiintrochatgpt} and DeepSeek \citep{guo2025deepseek} have been trained on vast amounts of human data and are finely aligned with human knowledge through techniques like reinforcement learning from human feedback \citep{bai2022training}, endowing them with a breadth and depth of knowledge that surpasses most individuals. Nevertheless, the molecular knowledge acquired by LLMs intuitively follows a long-tail distribution. For well-studied molecular properties, LLMs may have accumulated sufficient experience, but for less-explored areas, they may lack adequate reference rules. Furthermore, due to the well-known phenomenon of hallucination, LLMs often provide seemingly plausible answers even when they lack sufficient knowledge. While LLM4SD offers an effective means of utilizing the knowledge embedded in LLMs, it cannot fully eliminate hallucinations and biases. Therefore, we argue that integrating molecular structural information—particularly representations learned from pre-trained structural models—with human prior knowledge represents a highly promising direction for advancing MPP.

In this work, we propose the integration of knowledge from LLMs with molecular structural information to enhance MPP. Similar to LLM4SD, we extract prior knowledge from LLMs based on different types of molecular properties and further prompt LLMs to infer potential knowledge by providing molecular samples related to the target properties. We instruct the LLMs to generate both relevant knowledge and executable function code, which are then used to vectorize the molecules and obtain knowledge-based molecular features. These extracted knowledge features are subsequently fused with structural features obtained from pre-trained molecular structure models. By combining knowledge features with structural features, our model not only leverages the breadth of human expertise but also learns direct mappings between structure and properties from the structural features. We employ three state-of-the-art LLMs for knowledge extraction, including GPT-4o, GPT-4.1 \citep{openaiintrochatgpt}, and DeepSeek-R1 \citep{guo2025deepseek}. Extensive comparative experiments demonstrate the effectiveness of our approach and confirm that LLMs can provide reliable knowledge for MPP.

\section{Related work}
\subsection{Expert-crafted feature based MPP.}
Many earlier MPP models were based on expert-crafted features, primarily including two categories: molecular descriptors and molecular fingerprints. Molecular descriptors are quantitative features or attributes extracted from molecular structures. These features can range from simple numerical values to complex mathematical expressions and are used to describe the physicochemical properties, topological structures, and electronic characteristics of molecules \citep{khan2016descriptors}. Molecular fingerprints, on the other hand, are binary vectors or bit strings used to represent molecular structures. They map molecular structures to a fixed-length bit string, where each bit typically indicates the presence or absence of a specific substructure or chemical feature \citep{barnard1997chemical, sheridan1996chemical, DANISHUDDIN20161291}. The performance of such models often depends on the selection and quality of these expert features \citep{ZHANG201756, 10.1039/c7tx00259a}. These expert-crafted features are commonly used to train traditional machine learning models, such as LightGBM, random forest, and support vector machines \citep{zhang2019lightgbm, zhang2017carcinopred}. Additionally, some studies have used molecular fingerprints as input features to train more modern deep neural network architectures, such as BERT \citep{wen2022fingerprints} and GNN \citep{cai2022fp}.

\subsection{Graph-based MPP}
Molecules can naturally be viewed as graph structures, where atoms are considered as nodes and covalent bonds between atoms as edges. With the advancement of graph neural networks (GNNs), many studies have shifted towards using GNNs for MPP tasks. However, training deep neural networks requires a substantial amount of high-quality labeled data. Due to the scarcity of labeled molecular data, previous work has explored various methods to enhance model prediction capabilities, such as pre-training, contrastive learning, and knowledge augmentation. Pre-training involves constructing self-supervised learning tasks and training on large unlabeled datasets to obtain better initial weights, thereby improving the model's predictive performance on downstream tasks. Common self-supervised paradigms include node attribute reconstruction, context learning, and constructing multi-view graphs for contrastive learning \citep{velivckovic2017graph, you2020graph, xia2023mole}. Contrastive learning methods enhance model generalization by aligning different modalities of molecules through contrastive loss, and can also be considered a form of pre-training task \citep{liu2021pre, kuang20243d, you2020graph, xiang2023chemical, wang2022molecular}. Knowledge augmentation aims to improve the model's understanding of molecules by leveraging external knowledge bases. For instance, KPGT \citep{li2023knowledge} utilizes additional knowledge as molecular labels, enabling the model to effectively capture both structural and semantic information within molecular graphs.

\subsection{LLM-based MPP}
With the significant success of LLMs in general domains, recent efforts have been made to leverage language models (LM) or LLMs for MPP. For instance, \cite{liu-etal-2023-molxpt} replaced the CHEML IDs of molecules in the training corpus with SMILES sequences to create text data encapsulating SMILES, and then fine-tuned MolT5 \cite{edwards-etal-2022-translation} to predict molecular properties and translate between molecules and natural language. Similarly, \cite{liu2024moleculargpt} developed a mixed instruction set based on SMILES to fine-tune LLaMA, achieving notable performance in few-shot and zero-shot MPP. MolFM \citep{luo2023molfm} trained a multimodal foundational model for molecules by integrating knowledge graphs, molecular structures, and natural language. BioT5 \citep{pei-etal-2023-biot5} jointly trained on SELFIES of molecules, protein sequences, and natural language, also enabling MPP and related tasks.

However, these approaches require extensive fine-tuning of the language models to learn the mapping between molecules and natural language from large datasets, rather than directly extracting human knowledge about molecules from pre-trained LMs. Existing general LLMs, such as ChatGPTs \citep{openaiintrochatgpt} and DeepSeeks \citep{guo2025deepseek}, have been thoroughly trained on vast human knowledge bases, suggesting they may have internalized human knowledge and experience regarding molecules. LLM4Mol \citep{qian2023can} annotates molecular sequences using ChatGPT and employs the embeddings of the annotated text for downstream tasks. LLM4SD \citep{zheng2025large} extracts task-related rules using LLMs, vectorizes molecules based on these rules, and then trains a random forest model for MPP. Both methods utilize the knowledge LLMs have acquired from human experience but overlook the structural information inherent in molecules.

Intuitively, the molecular knowledge learned by LLMs from human corpora follows a long-tail distribution. For well-researched molecular properties, LLMs may have acquired sufficient experience, whereas for knowledge lacking extensive research literature, LLMs might not provide adequate reference rules. Therefore, incorporating the intrinsic structure of molecules is essential for molecular-related tasks. This work proposes the first method for MPP that combines LLM-driven knowledge with molecular structure.

\section{Method}
For each MPP task \( t \in T \), we have a dataset \( D_t = (X_t, Y_t) \), where each \( x_{t,i} \in X_t \) is a molecule's SMILES notation, and each molecule corresponds to a molecular graph \( g_{t,i} \in G_t \). The true label for each molecule is \( y_{t,i} \in Y_t \). 

\begin{figure}[t]
  \centering
  \includegraphics[width=0.45\textwidth]{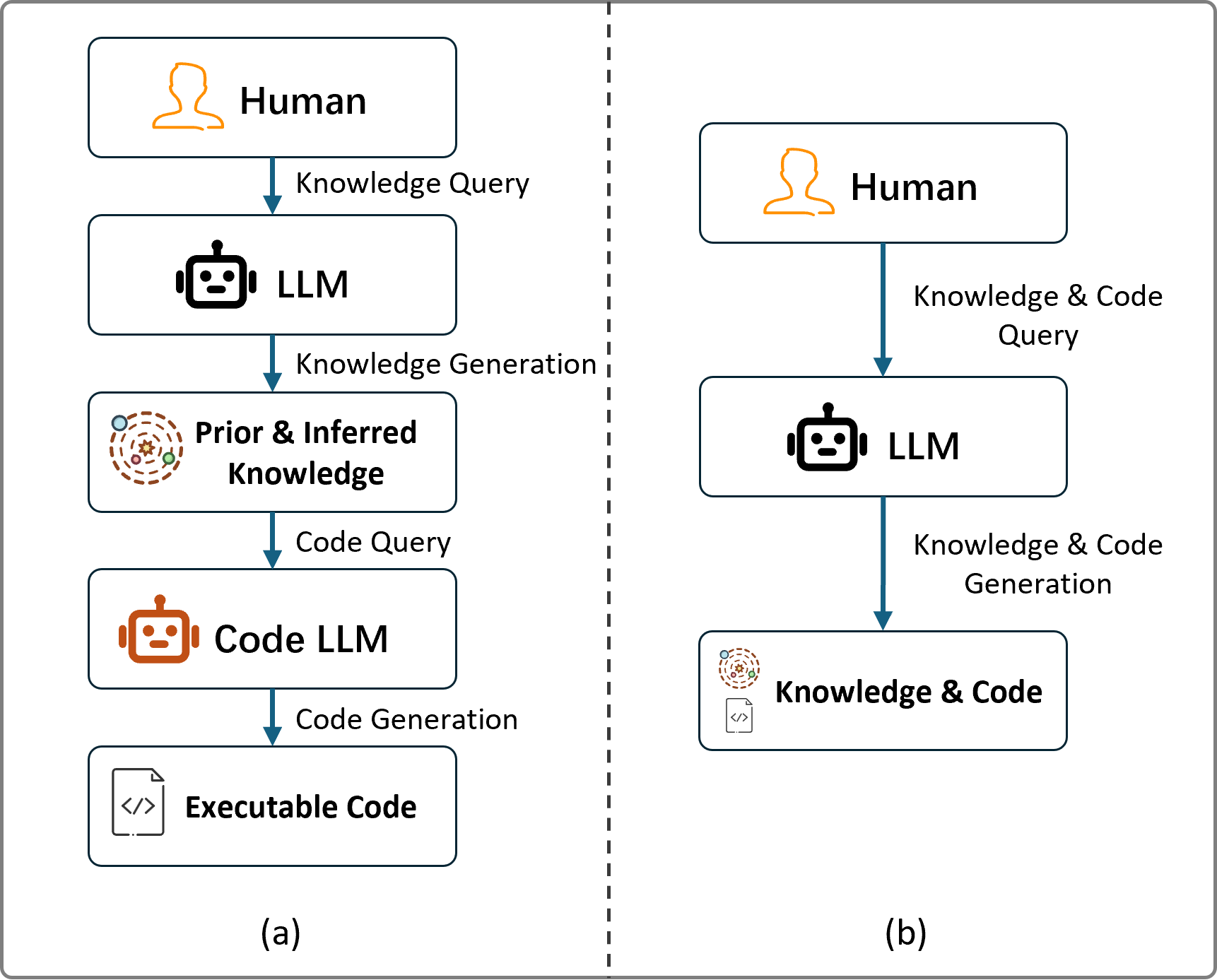}
  \caption{We draw on the LLM4SD approach to extract knowledge from LLMs. However, unlike LLM4SD, which separates the process into two distinct phases—knowledge extraction and code generation (a), we optimize the prompt to simultaneously generate both the knowledge and the associated executable functions in a single step (b).}
  \label{fig:knowledge}
\end{figure}

We leverage the knowledge provided by LLMs as molecular knowledge features. Drawing inspiration from LLM4SD, we extract both prior knowledge and inference knowledge. Prior knowledge refers to information that LLMs have acquired from extensive human literature, while inference knowledge is derived from the LLMs’ reasoning capabilities, enabling them to infer knowledge from a limited number of samples. LLM4SD implements knowledge extraction in a segmented manner: it first assigns the LLM the role of a chemistry expert to summarize a set of judgment rules based on the task type or to infer possible rules from given molecular samples. These rules are then passed to another code LLM, which generates executable code accordingly. We optimize this process by designing prompts that instruct the LLM to directly generate both the rules and the corresponding executable function code. Figures \ref{fig:knowledge}-(a) and (b) illustrate the differences between our knowledge extraction method and that of LLM4SD. For a given task \( t \) and molecule \( x_{t,i} \), we dynamically construct the molecular knowledge representation \( v_{t,i}^{K} \) using a prior knowledge function \( f_t^{P}(\cdot) \) and an inference knowledge function \( f_t^{I}(\cdot) \). The extracted molecular knowledge representations are then fused with the molecular graph representation \( f_G(g_{t,i}) \), and the combined features serve as the input to the model.


\begin{algorithm}[tb]
\caption{Knowledge and Structure Fusion}
\label{alg1}
\textbf{Input}: Task $t$, molecular sequence $x_{t,i}$, molecular graph $g_{t,i}$, LLM, task prompt $\mathcal{P}_t^P, \mathcal{P}_t^I$.\\
\textbf{Output}: $h_{t,i}$
\begin{algorithmic}[1] 
    \STATE $f_t^{P} = \mathrm{LLM}(t, \mathcal{P}_t^P)$
    \STATE $f_t^{I} = \mathrm{LLM}(t, \mathcal{P}_t^I)$
    \STATE $v_{t,i}^{K} = \operatorname{concat}(f_t^{P}(x_{t,i}), f_t^{I}(x_{t,i}))$
    \STATE $h_{t,i}^{K} = \mathrm{fc}(\sigma(\mathrm{fc}(\mathrm{BN}(v_{t,i}^{K}))))$
    \STATE $h_{t,i}^{S} = f_G(g_{t,i}) = \mathrm{GIN}_{\text{pre-trained}}(g_{t,i})$
    \STATE $h_{t,i} = \mathrm{fc}(\sigma(\mathrm{fc}(\operatorname{concat}(h_{t,i}^{K}, h_{t,i}^{S}))))$
    \STATE \textbf{return} $h_{t,i}$
\end{algorithmic}
\end{algorithm}




Mutual information quantifies the degree of interdependence between two random variables. Specifically, it measures the reduction in uncertainty of one variable given knowledge of the other. Mathematically, mutual information is defined as:
$$
I(X;Y) = H(X) + H(Y) - H(X,Y).
$$

Here, \( H(X) \) and \( H(Y) \) denote the entropies of the random variables \( X \) and \( Y \), respectively, while \( H(X,Y) \) represents their joint entropy. Maximizing mutual information involves leveraging \( Y \) to provide as much information as possible to reduce the uncertainty of \( X \). This approach has been widely validated as effective in areas such as contrastive learning, multimodal fusion, and self-supervised learning \citep{you2020graph, han-etal-2021-improving, ben2023reverse}. In this work, the knowledge vector and structure vector of molecules are regarded as two distinct modalities. Our objective is to introduce knowledge information to reduce the uncertainty of predictions based solely on graph structure. Therefore, we seek to better extract complementary information between modalities by maximizing the mutual information between the knowledge latent representation and the structure latent representation. To this end, we adopt the MINE \citep{belghazi2018mutual} approach, which utilizes neural networks to maximize mutual information in continuous space:

\begin{align*}
    \text{argmax}_{\theta}I(H^{K},H^{S}) = \text{argmax}_{\theta} \mathbb{E}_{p(H^{K},H^{S})}[T_\theta] \\
        - \log(\mathbb{E}_{p(H^{K})\cdot p(H^{S})}[e^{T_\theta}] ),
\end{align*}
where \( T \) represents a neural network, and \( \theta \) are the parameters of \( T \).

\begin{table*}[htbp]
  \centering
  \small
    \begin{tabular}{c|cccc|cccc}
    \toprule
    \multirow{2}[4]{*}{\textbf{method}} & \multicolumn{4}{c|}{\textbf{ROC\_AUC}} & \multicolumn{4}{c}{\textbf{PRECISION}} \\
\cmidrule{2-9}          & \textbf{HIV} & \textbf{BACE} & \textbf{CinTox} & \textbf{BBBP} & \textbf{HIV} & \textbf{BACE} & \textbf{CinTox} & \textbf{BBBP} \\
    \midrule
    GraphMVP & 77.73  & 82.71  & 75.89  & 74.83  & 41.43  & 83.83  & 24.02  & 61.18  \\
    GraphCL & 74.16  & 77.68  & 72.61  & 74.08  & 37.76  & 83.04  & 27.39  & 60.13  \\
    InfoGraph & 74.86  & 80.27  & 81.15  & 74.60  & 33.33  & 81.42  & 19.41  & 63.01  \\
    AttrMask & 73.38  & 83.94  & 76.76  & 74.08  & 30.65  & 87.15  & 24.86  & 61.29  \\
    ContextPred & 74.53  & 86.18  & 72.72  & 74.91  & 18.42  & 86.27  & 22.04  & 56.97  \\
    EdgePred & 74.28  & 77.97  & 62.59  & 70.91  & 41.18  & 80.68  & 7.08  & 60.53  \\
    GPT-GNN & 76.68  & 84.68  & 60.83  & 69.79  & 33.06  & 85.85  & 4.76  & 54.80  \\
    G-Motif & 77.15  & 81.59  & 72.32  & 75.02  & 46.43  & 82.44  & 16.69  & 60.13  \\
    \midrule
    GraphMVP (+GPT-4o) & +2.10 & +0.33 & +0.93 & +1.74 & +2.86 & +0.65 & -6.88 & -0.71 \\
    GraphCL(+GPT-4o) & +4.09 & +2.29 & +2.32 & +1.0  & +15.82 & -1.62 & -11.47 & +0.45 \\
    InfoGraph (+GPT-4o) & +4.79 & +1.97 & +1.21 & +2.12 & +24.67 & +3.06 & +19.29 & +0.15 \\
    AttrMask (+GPT-4o) & +6.33 & +1.37 & +5.46 & +0.49 & +28.28 & +1.33 & +12.14 & -0.2 \\
    ContextPred (+GPT-4o) & +3.81 & +1.66 & -4.13 & +0.12 & +31.58 & +1.49 & +12.25 & -0.31 \\
    EdgePred (+GPT-4o) & +3.59 & +5.34 & +7.93 & +3.61 & +8.82 & +0.72 & +1.62 & +1.79 \\
    GPT-TNN (+GPT-4o) & +2.95 & +0.76 & -0.14 & +1.82 & +14.48 & -0.32 & +1.40 & +4.04 \\
    G-Motif (+GPT-4o) & +1.64 & +1.33 & -6.71 & +0.2  & -0.16 & +1.45 & +6.42 & +2.1 \\
    \midrule
    GraphMVP (+GPT-4.1) & +1.72 & +0.98 & +4.95 & +1.72 & +2.80 & +1.45 & -7.18 & -0.52 \\
    GraphCL (+GPT-4.1) & +4.73 & +3.38 & +7.56 & +0.89 & +24.15 & -1.01 & +19.15 & +1.33 \\
    InfoGraph (+GPT-4.1) & +4.80 & +1.99 & +3.24 & +2.07 & +17.58 & +3.19 & +3.06 & +1.09 \\
    AttrMask (+GPT-4.1) & +7.23 & +2.48 & +0.37 & +1.75 & +22.55 & +0.31 & +6.18 & +0.03 \\
    ContextPred (+GPT-4.1) & +6.79 & +1.35 & -2.71 & +0.03 & +49.32 & +1.44 & +13.02 & +1.67 \\
    EdgePred (+GPT-4.1) & +5.39 & +4.14 & +9.41 & +3.94 & +23.53 & +1.78 & +2.97 & +0.65 \\
    GPT-TNN (+GPT-4.1) & +3.03 & +1.96 & +2.92 & +2.15 & +15.55 & +1.62 & +14.89 & +3.48 \\
    G-Motif (+GPT-4.1) & +2.66 & +2.65 & -2.49 & +0.07 & +12.55 & +2.88 & +5.48 & +1.14 \\
    \midrule
    GraphMVP (+DeepSeek-r1) & +0.54 & -0.01 & +0.36 & +0.94 & +2.10 & +1.21 & -6.03 & -0.83 \\
    GraphCL (+DeepSeek-r1) & +4.64 & +2.92 & +0.80 & +1.05 & +18.85 & +0.68 & +4.94 & +1.94 \\
    InfoGraph (+DeepSeek-r1) & +4.54 & +1.18 & +1.13 & +2.38 & +11.11 & +2.27 & +21.60 & -0.23 \\
    AttrMask (+DeepSeek-r1) & +6.31 & +1.65 & -5.10 & +0.31 & +29.77 & -1.51 & -10.69 & -1.78 \\
    ContextPred (+DeepSeek-r1) & +3.50 & +0.53 & -1.38 & +0.3  & +42.98 & +0.31 & +4.93 & -0.07 \\
    EdgePred (+DeepSeek-r1) & +5.62 & +2.3  & +7.43 & +3.67 & +12.02 & +2.23 & +3.68 & +3.69 \\
    GPT-TNN (+DeepSeek-r1) & +2.90 & -0.07 & +7.23 & +2.29 & +22.71 & -1.35 & +3.70 & +2.89 \\
    G-Motif (+DeepSeek-r1) & +2.60 & +1.77 & +4.05 & +0.74 & -9.84 & +2.28 & +8.73 & +1.45 \\
    \midrule
    GraphMVP (GNN+Desc.) & N/A   & +0.64 & -18.04 & -1.31 & N/A   & +5.93 & -6.48 & -3.31 \\
    GraphCL (GNN+Desc.) & N/A   & +1.45 & -20.40 & -2.7  & N/A   & +0.41 & -10.92 & -1.03 \\
    InfoGraph (GNN+Desc.) & N/A   & +0.33 & -19.27 & -0.99 & N/A   & +1.11 & -7.04 & +0.24 \\
    AttrMask (GNN+Desc.) & N/A   & +1.22 & -20.49 & -4.15 & N/A   & +2.57 & -9.79 & -2.18 \\
    ContextPred (GNN+Desc.) & N/A   & +1.03 & -18.71 & -4.74 & N/A   & +4.84 & -6.36 & -1.53 \\
    EdgePred (GNN+Desc.) & N/A   & +2.39 & -4.72 & +0.53 & N/A   & +5.69 & +4.30 & +1.86 \\
    GPT-TNN (GNN+Desc.) & N/A   & -0.38 & +6.91 & +0.34 & N/A   & +2.85 & +2.42 & +2.43 \\
    G-Motif (GNN+Desc.) & N/A   & +0.96 & -9.47 & -1.05 & N/A   & +3.61 & -12.76 & -0.15 \\
    \bottomrule
    \end{tabular}%
    \caption{The experimental results for integrating knowledge from different LLMs using various pre-trained models as backbones are presented. In this context, "GNN+Desc." denotes the unselective integration of all descriptors. The symbol ‘+’ indicates an improvement in model performance following the incorporation of additional knowledge, whereas ‘-’ signifies a decrease in performance. Notably, for the HIV dataset, the descriptor features are excessively sparse and highly similar, which leads to vanishing gradients within the model and causes GNN+Desc. to fail in making accurate predictions. We ran each experiment five times using five different random seeds and reported the average results.}
  \label{tab:main_results}%
\end{table*}%

\begin{figure*}[t]
  \centering
  \includegraphics[width=1\textwidth]{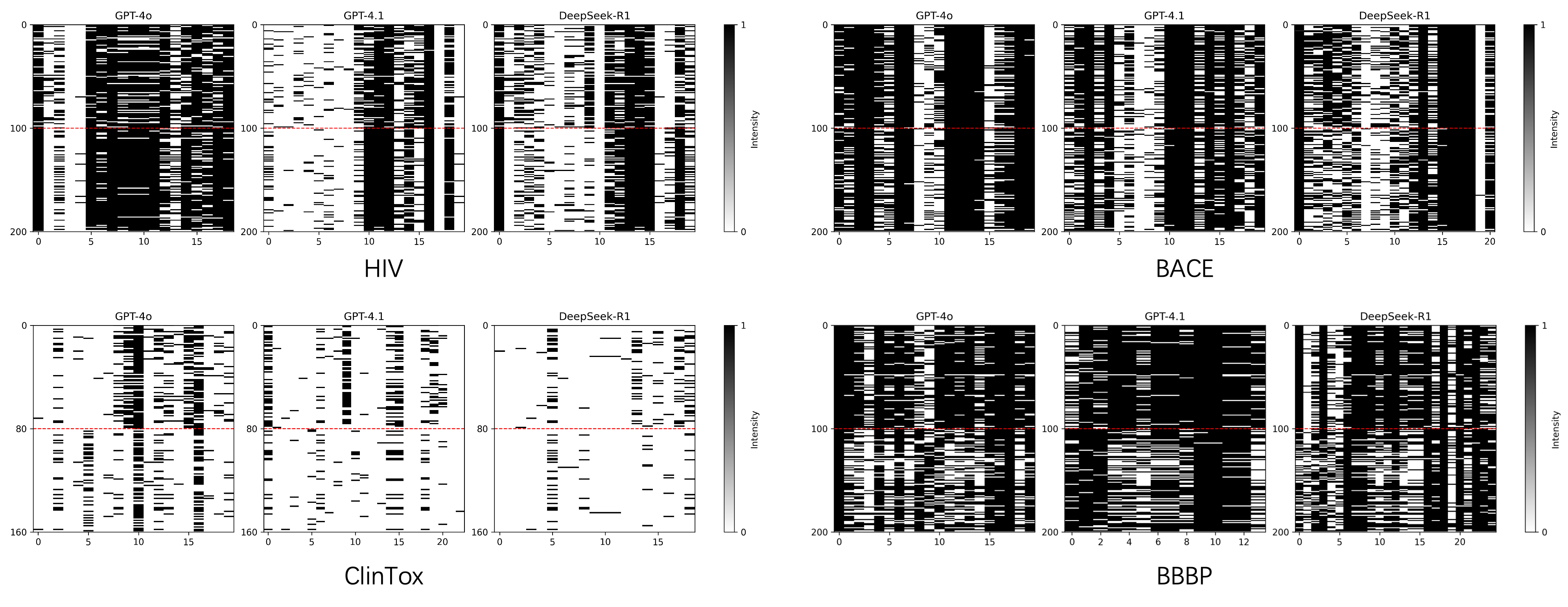}
  \caption{Visualization of knowledge vectors generated by different LLMs on each dataset. Each row in a subfigure represents a molecule, and each column corresponds to a feature. The top 100 rows are randomly selected positive samples, while the bottom 100 rows are randomly selected negative samples.}
  \label{fig:colormap_bbbp}
\end{figure*}

\section{Experiments} 
\textbf{Datasets.} The HIV dataset, derived from the Drug Therapeutics Program AIDS Antiviral Screen, contains results from testing over 40,000 compounds for their ability to inhibit HIV replication, with outcomes categorized as confirmed inactive, active, or moderately active. The BACE dataset focuses on human beta-secretase 1 inhibitors, providing both quantitative IC50 values and qualitative binary labels for small molecule inhibitors across a broad affinity range. It includes 154 inhibitors for affinity prediction, 20 for pose prediction, and 34 for free energy prediction. The ClinTox dataset contrasts FDA-approved drugs with those that failed clinical trials due to toxicity, featuring classification tasks for 1,491 compounds based on clinical trial toxicity and FDA approval status. The BBBP dataset, originating from a study on blood-brain barrier permeability, offers binary labels indicating whether compounds can penetrate the BBB. This dataset supports the development of machine learning models for predicting BBB permeability, which is crucial for drug discovery and the design of neurological therapeutics \citep{wu2018moleculenet}.


\textbf{Baselines.} Since our goal is to enhance the capabilities of pre-trained molecular structure models through knowledge integration, we selected several classic pre-training models as baselines and backbones. These include GraphMVP \citep{liu2021pre}, GraphCL \citep{you2020graph}, InfoGraph \citep{sun2019infograph}, AttrMask \citep{hu2019strategies}, ContextPred \citep{hu2019strategies}, EdgePred \citep{hamilton2017inductive}, GPT-GNN \citep{hu2020gpt}, and GROVER \citep{rong2020self}. All comparative experiments were conducted using the same molecular scaffold dataset split, with model parameters set to the default values provided by the respective authors. Furthermore, we observed that the extracted knowledge primarily relies on the Descriptors modules of RDKit \citep{landrum2013rdkit} for molecular descriptor extraction. Given that deep neural networks are capable of capturing nonlinear relationships directly from complex data, we integrated all descriptors provided by LLMs, removed any thresholds, and directly used the raw descriptor features in combination with structural features. This approach was designed to assess whether LLMs possess inherent feature selection capabilities. We refer to this control group as GNN+Desc, and its model architecture and training parameters are identical to those of the experimental groups.

\textbf{Metrics.} Many studies on MPP evaluate models solely using the Receiver Operating Characteristic Area Under the Curve (ROC AUC) \citep{fangmol, liu2021pre, wang2022molecular}. This metric considers recall and false positive rates across various thresholds, providing a comprehensive assessment of model performance. However, in highly imbalanced datasets, ROC AUC can yield overly optimistic evaluations. For example, in the ClinTox dataset, the ratio of positive to negative samples is 1:12.19, rendering ROC AUC insufficient as a sole metric. Therefore, in addition to ROC AUC, we also employ precision as an evaluation metric. Precision measures the proportion of true positive samples among those predicted as positive, offering more meaningful insights in scenarios with imbalanced class distributions.

\textbf{Knowledge Extraction.} We extract both prior knowledge and inference knowledge for each task from three mainstream LLMs: GPT-4o, GPT-4.1, and DeepSeek-R1. The prompts used for this process are detailed in Appendix 1. For inference knowledge that requires sample data, we randomly select 50 positive samples from each training set, using the same samples across all LLMs to ensure consistency.

\subsection{Experimental result}
The overall experimental data are presented in Table \ref{tab:main_results}. For a more intuitive comparison, these data are also visualized in Figures 7, 8, 9, and 10 in Appendix. Based on the experimental results, we can draw the following conclusions:

\textbf{Integrating knowledge extracted from LLMs can effectively enhance the performance of MPP.} Although different pre-training strategies may lead models to different local optima, incorporating knowledge from LLMs almost always results in performance improvements. Taking the knowledge provided by GPT-4.1 as an example, the most significant improvement was observed on the ClinTox dataset, where the ROC AUC increased by up to 9.41\%, and an average increase of 2.91\%. On the HIV, BACE, and BBBP datasets, improvements were also observed, with average increases of 4.54\%, 2.37\%, and 1.58\%, respectively. We also observed that, under certain pre-training strategies, integrating knowledge could occasionally lead to negative gains. For example, on the BBBP dataset, the AM method experienced a 0.11\% decrease in ROC AUC after the integration of knowledge from DeepSeek-R1. 



\textbf{The knowledge provided by different LLMs has varying impacts on model performance.} Overall, GPT-4.1 achieved the highest average improvement across all four datasets, with increases of 4.54\%, 2.37\%, 2.91\%, and 1.58\%, respectively. On the HIV, ClinTox, and BBBP datasets, DeepSeek-R1 provided a greater average improvement than GPT-4o, whereas on the BACE dataset, GPT-4o yielded a larger gain.

\textbf{Directly integrating all molecular descriptors does not necessarily lead to improved performance.} Although the direct fusion of all descriptors resulted in an average increase of 0.96\% on the BACE dataset in ROC AUC, it actually led to a decrease in ROC AUC performance on the ClinTox and BBBP datasets. Specifically, on the ClinTox dataset, the GNN+Desc. approach resulted in an average 1.76\% decrease in ROC AUC, which is 13.88\%, 15.93\%, and 14.84\% lower than the results achieved by GPT-4o, GPT-4.1, and DeepSeek-R1, respectively. On the BBBP datasets, GNN+Desc. also exhibited average decreases of 1.76\% compared to the baseline models. These results demonstrate that leveraging prior knowledge provided by LLMs is indeed more effective than simply combining all descriptors in a brute-force manner. We also observed that, on the HIV dataset, incorporating all descriptors led to features that were excessively sparse and highly similar, causing the model to suffer from vanishing gradients and fail to generate accurate predictions. We will analyze these features in detail in the next section.

\subsection{Knowledge Analysis}
\textbf{Knowledge visualization.} In addition, we visualized the knowledge extracted from different LLMs. Figure \ref{fig:colormap_bbbp} presents the visualization results of knowledge features from different LLMs on each dataset. In each subfigure, each row represents a molecule, and each column represents a feature dimension. The first 100 rows correspond to 100 randomly selected positive samples, while the last 100 rows correspond to 100 randomly selected negative samples. We can observe that for the HIV, ClinTox, and BBBP datasets, the knowledge provided by each LLM exhibit clear and visually discernible discriminative power between positive and negative samples. However, for the BACE dataset, these knowledge vectors do not display any obvious visual discriminative ability. Therefore, in such cases, it is essential to incorporate molecular structural information.

\begin{figure}[htbp]
  \centering
  \includegraphics[width=0.48\textwidth]{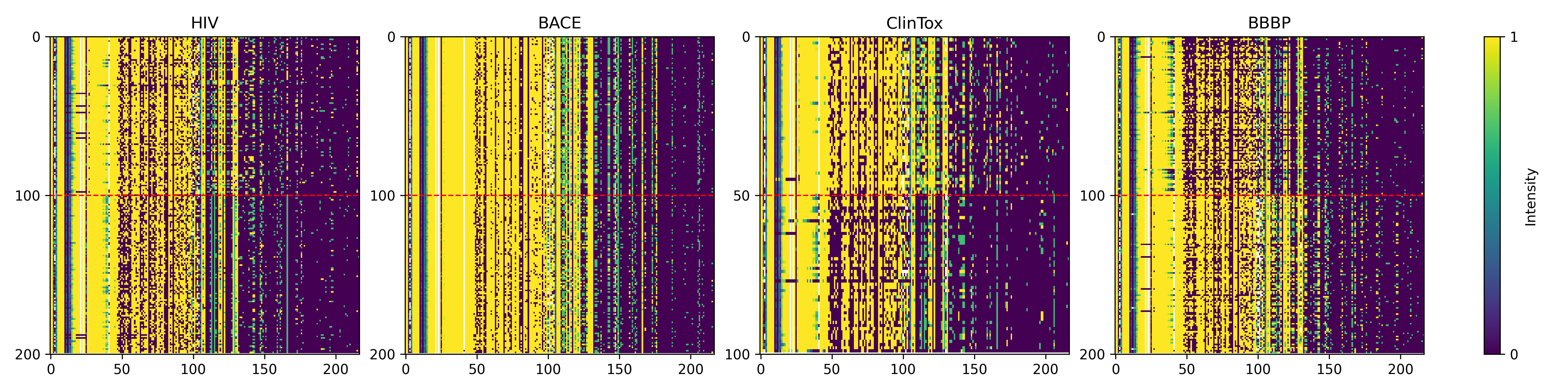}
  \caption{Visualization of vectors using all molecular descriptors. From left to right, the datasets are HIV, BACE, ClinTox, and BBBP.}
  \label{fig:colormap_desc}
\end{figure}


\begin{figure*}[t]
  \centering
  \includegraphics[width=1\textwidth]{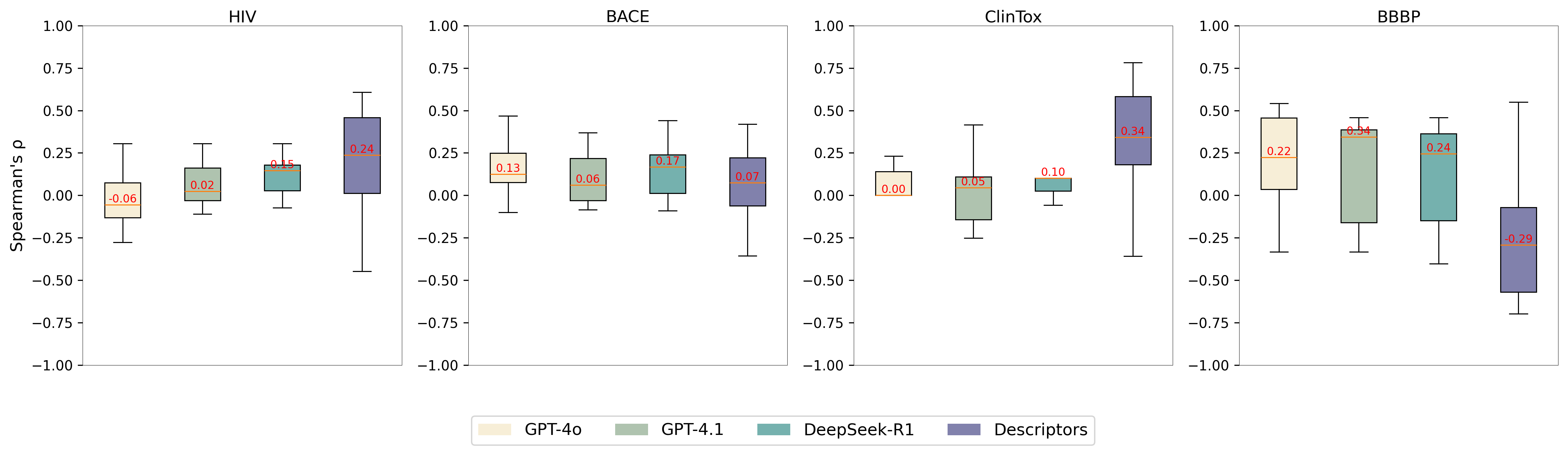}
  \caption{Prior Knowledge and the Spearman Rank Correlation Coefficient of Molecular Properties.}
  \label{fig:spearman_prior}
\end{figure*}

For comparison, we also visualized all descriptor features on the four datasets using the same settings. As shown in Figure \ref{fig:colormap_desc}, the results for HIV, BACE, ClinTox, and BBBP are displayed from left to right. It can be observed that, for ClinTox and BBBP, these descriptors also exhibit some visually discernible differences between positive and negative samples, but the differences are less pronounced than those in the knowledge vectors. Moreover, the descriptor vectors are much longer than the knowledge vectors provided by the LLMs, making it more challenging for the model to directly learn specific patterns from all the descriptors.

\textbf{Correlation between knowledge and molecular properties.} We separately analyzed the correlations between prior knowledge and inference knowledge generated by LLMs and molecular properties, using the Spearman rank correlation coefficient \citep{sedgwick2014spearman}. For prior knowledge, we compared the correlations between LLM-generated prior knowledge and molecular properties with those between all descriptors and molecular properties. As shown in Figure \ref{fig:spearman_prior}, the correlations for prior knowledge vary considerably across different datasets. On the BBBP dataset, prior knowledge generated by all three LLMs shows a high correlation with molecular properties, outperforming native descriptors. On the BACE dataset, the correlation of prior knowledge is also satisfactory. However, on the HIV and ClinTox datasets, the overall correlation between LLM-generated prior knowledge and molecular properties is significantly lower than that of the native descriptors.

For inference knowledge, as shown in Figure 6 in Appendix, the correlation with molecular properties is higher on the BBBP and ClinTox datasets compared to HIV and BACE. Notably, on BBBP and ClinTox, some inference knowledge achieves a correlation greater than 0.5, indicating a strong association. Overall, the inference knowledge generated by GPT-4o demonstrates a relatively high correlation.


\textbf{Knowledge redundancy and conflict.} When extracting prior and inferential knowledge, we explicitly require each LLM to provide at least 10 rules. For example, in the ClinTox task, our prompts for extracting prior and inferential knowledge are as follows: “How can one determine if a molecule has clinical toxicity based on its SMILES? Please provide at least 10 computational rules for assessment and a Python-based computational method. Each rule should be implemented as a function that takes a molecule as input and returns a 0/1 value.” and “How can one determine if a molecule is toxic based on its SMILES representation? Please infer possible criteria from the clinically validated toxic SMILES provided below, and offer at least 10 computational methods using Python. Each rule should be implemented as a function, with the input being a molecule and the output being a 0/1 value. Below are the molecules that have been clinically validated as toxic:......”

\renewcommand{\arraystretch}{0.9} 
\begin{table}[htbp]
  \centering
  \small 
  \begin{tabularx}{0.48\textwidth}{@{}c|c@{}c|c@{}c|c@{}c@{}}
    \toprule
    \multirow{2}{*}{\textbf{DATASET}} & \multicolumn{2}{c|}{\textbf{GPT-4o}} & \multicolumn{2}{c|}{\textbf{GPT-4.1}} & \multicolumn{2}{c}{\textbf{DeepSeek-R1}} \\
    \cmidrule{2-7}
    & \textbf{repeat} & \textbf{clash} & \textbf{repeat} & \textbf{clash} & \textbf{repeat} & \textbf{clash} \\
    \midrule
    HIV   & 2/20  & 6/20  & 0/20 & 2/20 & 1/20  & 3/20 \\
    BACE  & 1/20  & 1/20  & 0/20 & 2/20 & 0/20  & 6/20 \\
    BBBP  & 8/20  & 4/20  & 1/20 & 6/20 & 3/20  & 2/20 \\
    ClinTox & 1/20  & 2/20  & 0/25 & 4/25 & 2/20  & 2/20 \\
    \midrule
    AVG   & 15.00\% & 16.25\% & 1.25\% & 17.50\% & 7.50\% & 16.25\% \\
    \bottomrule
  \end{tabularx}
  \caption{The repetition and conflict rates of knowledge provided by different LLMs.}
  \label{tab:conflict}
\end{table}

In most cases, each LLM generates 10 rules, except for the ClinTox dataset, where GPT-4.1 generated 15 rules. We also observed that the generated prior and inferential knowledge sometimes exhibit repetition and conflict. Repetition refers to two rules having identical evaluation criteria, while conflict refers to two rules addressing the same aspect but providing different thresholds. For instance, in the HIV dataset, both the prior and inferential knowledge provided by GPT-4o include the same rule: “has aromatic rings,” with identical computational methods. However, when assessing the topological polar surface area (TPSA) of molecules, the prior knowledge suggests that molecules may be toxic when TPSA $<$ 140, whereas the inferential knowledge suggests toxicity when 20 $\leq$ TPSA $\leq$ 200.

The repetition and conflict of knowledge across various datasets are detailed in Table \ref{tab:conflict}. For GPT-4o, the average repetition rate is 15\% and the conflict rate is 10\%; for GPT-4.1, the average repetition rate is 1.25\% and the conflict rate is 16.25\%; and for DeepSeek-R1, the average repetition rate is 3.75\% and the conflict rate is 10\%.

\subsection{Conclusion}

We enhance molecular representation by integrating knowledge vectors from LLMs with structural features, addressing the limited discriminative power of these vectors for certain properties. This approach, tested across multiple pretrained models, proves effective, highlighting the potential for applying this fusion of LLM-derived knowledge and traditional features to other fields, offering new insights and advancements.



\newpage
\bibliography{aaai2026}

\end{document}